\documentclass[lettersize,journal]{IEEEtran}
\pdfoutput=1
\usepackage{amsmath,amsfonts}
\usepackage{algorithmic}
\usepackage{algorithm}
\usepackage{booktabs}
\usepackage{multirow}
\usepackage[table,xcdraw]{xcolor}
\usepackage{array}
\usepackage[caption=false,font=normalsize,labelfont=sf,textfont=sf]{subfig}
\usepackage{textcomp}
\usepackage{stfloats}
\usepackage{url}
\usepackage{verbatim}
\usepackage{graphicx}
\usepackage{cite}

\usepackage{bm}

\begin{document}

\title{Dynamic Frequency Modulation for Controllable Text-driven Image Generation}

\author{Tiandong Shi, Ling Zhao, Ji Qi, Jiayi Ma, Chengli Peng*

\thanks{
This work was supported by the National Natural Science Foundation of China (Grant Numbers 42301433 and 42271481) and the Natural Science Foundation of Hunan Province of China under Grant no.2024JJ6496. \textit{(Corresponding author: Chengli Peng.)}

Tiandong Shi, Ling Zhao, Chengli Peng are with the School of Geosciences and Info-Physics of Central South University, Changsha 410083, China. (e-mail: csushitd@csu.edu.cn; zhaoling@csu.edu.cn; pengcl@csu.edu.cn). Ji Qi is with the School of Geography and Remote Sensing, Guangzhou University, Guangzhou, 510006, China.(e-mail: jameschi95@foxmail.com). Jiayi Ma is with Electronic Information School, and also with the School of Robotics, Wuhan University, Wuhan 430072, China. (e-mail: jyma2010@gmail.com).
}
}

\maketitle

\begin{abstract}
The success of text-guided diffusion models has established a new image generation paradigm driven by the iterative refinement of text prompts. However, modifying the original text prompt to achieve the expected semantic adjustments often results in unintended global structure changes that disrupt user intent. Existing methods rely on empirical feature map selection for intervention, whose performance heavily depends on appropriate selection, leading to suboptimal stability. This paper tries to solve the aforementioned problem from a frequency perspective and analyzes the impact of the frequency spectrum of noisy latent variables on the hierarchical emergence of the structure framework and fine-grained textures during the generation process. We find that lower-frequency components are primarily responsible for establishing the structure framework in the early generation stage. Their influence diminishes over time, giving way to higher-frequency components that synthesize fine-grained textures. In light of this, we propose a training-free frequency modulation method utilizing a frequency-dependent weighting function with dynamic decay. This method maintains the structure framework consistency while permitting targeted semantic modifications. By directly manipulating the noisy latent variable, the proposed method avoids the empirical selection of internal feature maps. Extensive experiments demonstrate that the proposed method significantly outperforms current state-of-the-art methods, achieving an effective balance between preserving structure and enabling semantic updates.
\end{abstract}


\section{Introduction}
\IEEEPARstart{I}{n} recent years, text-guided diffusion models have achieved remarkable advancements in the fields of image synthesis \cite{DiT,Photorealistic} and have been widely used in diverse vision tasks \cite{tip_assess,tip_fusion1,tip_image_edit}, significantly surpassing previous generative models \cite{GAN,VAE} in terms of both generation fidelity and semantic alignment. This superior generative capability is fundamentally reshaping the landscape of human image generation. By replacing traditional workflows where artists manually create and modify content, these models have established a novel image generation paradigm driven by the iterative refinement of text prompts. Within this paradigm, representative models like Stable Diffusion \cite{LDM} assume the role of the creator, directly synthesizing images based on text prompts provided by the users. Consequently, users guide this creative process by iteratively refining prompts to ensure the generated images faithfully reflect their creative objectives.
\begin{figure*}[h]
\centerline{\includegraphics[width=1.\linewidth]{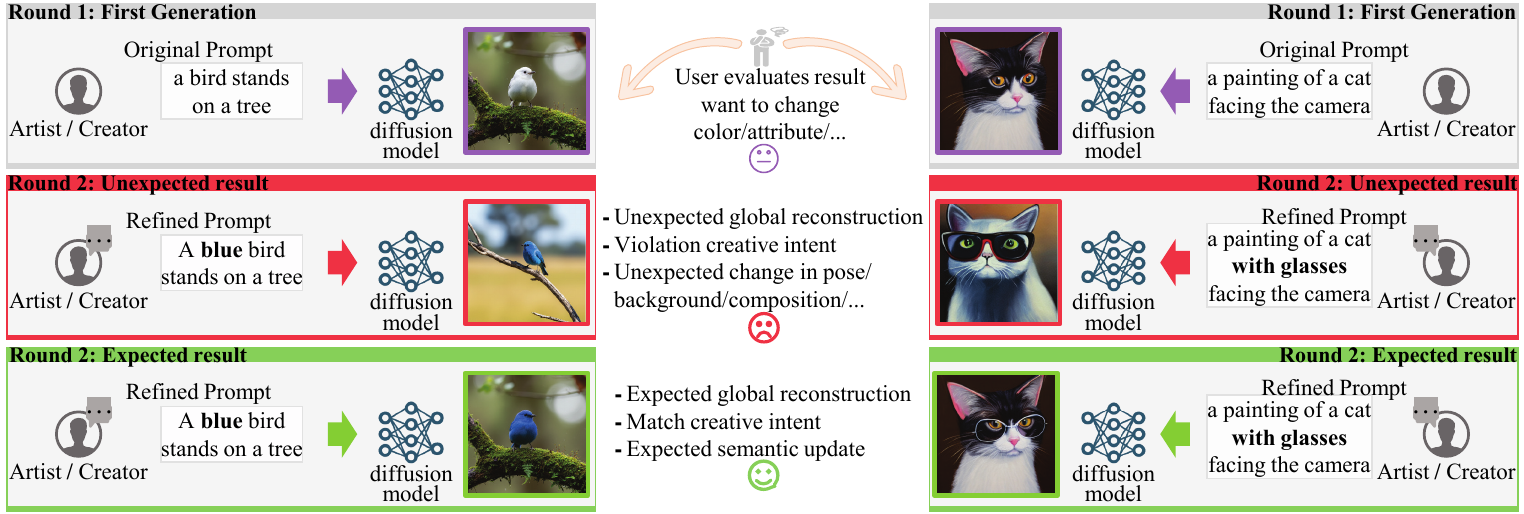}}
\caption{The key challenges faced by the image generation paradigm driven by the iterative refinement of text prompts. When the user changes the original text prompt for expected specific semantic adjustments, the newly generated images often exhibit unexpected reconstructions of composition, posture, background, etc., which goes against the creative intention.}
\label{fig1}
\end{figure*}

The effectiveness of the generation paradigm that relies on the iterative refinement of text prompts is grounded in a critical premise. The regenerated image should align with the specific adjustments made to the refined prompts, accurately reflecting the user’s intended meaning. However, current text-guided diffusion models demonstrate a significant sensitivity to variations in these prompts \cite{p2p}. Even minor modifications can lead to unintended alterations in the overall composition, rather than implementing the precise adjustments desired by the user. As illustrated in Fig.~\ref{fig1}, when a user refines the original prompt to change specific attributes such as color or pose, the resulting image often diverges significantly from the originally generated one, resulting in substantial changes in global composition, object positioning, and background elements. This kind of deviation fundamentally undermines the user's intent to make targeted adjustments. Thus, this heightened sensitivity to prompt changes severely limits the practicality of the image generation paradigm driven by iterative refinement of text prompts.

To address this challenge, recent methods propose to intervene on the spatial domain internal feature maps of denoising models \cite{pnp,FPE}. These spatial domain methods typically maintain structural consistency by transferring intermediate feature maps from the generation corresponding to the original prompt into the generation corresponding to the refined prompt. However, the selection of these feature maps is predominantly driven by empirical observation rather than explicit causal analysis. Lacking a rigorous theoretical foundation, such methods offer limited interpretability regarding the distinct mechanisms by which specific internal representations govern image structure versus style. Consequently, the generation outcomes suffer from inherent instability, and suboptimal feature maps transfer can cause the final output to diverge severely from the user's expectations.

This paper investigates the interpretable mechanisms governing the hierarchical emergence of image structure and fine-grained textures in text-guided diffusion models from an innovative frequency domain perspective. Through theoretical analysis and extensive experiments, we reveal that the lower-frequency components of the noisy latent variables first establish the structure framework in the early generation stage, and their influence weakened as the generation progresses, providing a basis for the transition to gradually synthesizing fine-grained textures by the higher-frequency components. Leveraging this insight, we propose a training-free Frequency Modulation Method (\emph{i.e.}, FMM). Unlike spatial-domain methods, the proposed method employs dynamic frequency spectrum intervention. Specifically, the FMM applies a frequency-dependent weighting function with dynamic decay to modulate the frequency components of the noisy latent variables corresponding to the refined text prompt with those corresponding to the original text prompt. This mechanism enforces global compositional consistency while simultaneously ensuring the synthesis of semantic content that aligns with the refined prompt. Crucially, this modulation employs a temporally adaptive relaxation strategy. As the generation progresses, it progressively relaxes the spectrum constraint, thereby facilitating the adaptive evolution of semantic content to achieve the targeted semantic adjustments. By operating directly on the noisy latent variables, the method circumvents the empirical selection of internal feature maps for spatial-domain methods, offering a more robust and interpretable solution.

Experimental results across two benchmarks demonstrate that our method strikes an optimal balance between preserving composition integrity and achieving precise semantic adjustments to reflect the refined prompt. Beyond the proposed method itself, this paper reveals the underlying spectrum mechanisms governing the hierarchical emergence of image composition and semantics, which offers a novel perspective for enhancing the controllability of the image generation paradigm driven by the iterative refinement of text prompts.

In summary, the main contributions of this paper are as follows:
\begin{itemize}
  \item We conduct theoretical analysis and experiments from a frequency-domain perspective to reveal the mechanism governing the hierarchical emergence of image structure and fine-grained textures in text-guided diffusion models.
  \item We reveal that lower-frequency components first determine the structure framework of the image during the early generation process, with their influence diminishing over time to allow for the progressive synthesis of fine-grained textures driven by higher-frequency components.
  \item We propose a novel training-free frequency modulation method that leverages a frequency-dependent weighting function with dynamic decay to incorporate spectrum constraints to preserve structure consistency, while dynamically relaxing the spectrum constraints to achieve semantic adjustment that responds to the refined prompt.
  \item Experiments across two benchmarks demonstrate that our method significantly outperforms the state-of-the-art methods. It achieves a superior balance between preserving composition integrity and achieving precise semantic adjustments to reflect the refined prompt.
\end{itemize}
\section{Related work}
In this paper, we classify the related work into two categories: optimization-based methods and training-free methods.
\subsection{Optimization-based Methods}
Based on their control conditions, this paper categorizes the related work into three types: text-instruction-based methods, methods that use a reference image together with a text prompt, and token-blending-based methods.

DiffusionCLIP \cite{diffusionclip}, InstructPix2Pix \cite{Instructpix2pix}, and Null-text Inversion \cite{null_text_inversion} are representative text-instruction-based methods. DiffusionCLIP is a text-guided image manipulation method that integrates a CLIP \cite{clip} loss to optimize the reverse process of the diffusion model. However, it requires a separate optimization for each text instruction, which limits its flexibility. InstructPix2Pix leverages data generated by a large language model and a text-guided diffusion model to enable instruction-based image editing through forward propagation. However, this forward-only method can fail to provide sufficient control when the instructions are ambiguous. Null-text Inversion achieves accurate inversion of an input image by optimizing the unconditional text embedding. However, it may perform poorly with context-specific instructions because it does not account for complex image details.

DreamInpainter \cite{Dreaminpainter} and SmartBrush \cite{Smartbrush} are representative methods that combine a reference image with a text prompt. DreamInpainter uses both inputs as constraints to generate content within a masked region that aligns with the subject of the reference image while adhering to the text description. The method balances the preservation of the subject’s identity with flexible semantic edits, but it struggles to produce coherent results when the reference image and text prompt convey conflicting information. SmartBrush employs text prompts to guide generation and incorporates a mask-precision controller, which accepts masks ranging from precise to blurred and thus allows flexible control over object generation. However,the reliance on masks may reduce flexibility in complex scenarios.

Token-binding-based methods are used to produce images featuring a specific subject or style. Representative methods include DreamBooth \cite{Dreambooth} and Textual Inversion \cite{An_image_one_world}. These methods usually require three to five reference images of the target subject or style to fine-tune the diffusion model or to learn a pseudo-word in the text-embedding space, thereby enabling the generation of personalized images that incorporate the target concept in diverse contexts. However, because they rely on the availability of sufficient reference images, their versatility in broader applications is limited.
\subsection{Training-free-based Methods}
Training-free methods manipulate generated images by directly leveraging internal features of diffusion models. These methods guide the generation process in the spatial domain using attention maps \cite{attention} or region masks, enabling localized edits based on text prompts without iterative optimization.

Attention-based methods leverage self-attention and cross-attention maps to guide localized image modifications based on text prompts. Representative methods include P2P \cite{p2p}, PnP \cite{pnp}, FPE \cite{FPE}, Pix2Pix-Zero \cite{pix2pix-zero}, MasaCtrl \cite{Masactrl}, and ShapeGuided Diffusion \cite{Shape_guided_diffusion}. P2P, PnP, and FPE analyze the attention layers of text-guided diffusion models and demonstrates that the attention maps capture pixel relationships. By inserting these maps corresponding to the reference text prompt into the generation controlled by the edited text prompt, it maintains the global structure while updating the content. Pix2PixZero automatically discovers edit direction vectors and applies cross-attention guidance to keep the structure consistent with the reference image during editing. In addition, TtfDf \cite{ttfdiffusion} discovers the decoupling direction that separates structure and semantics by performing gradient optimization in the h-space of the pre-trained model, thereby modifying the image semantics while preserving the structure. MasaCtrl combines query features from the edited image with key and value features from the reference image, preserving the latter’s appearance while allowing structure modifications specified by the edited text prompt. Shape-Guided Diffusion pairs text prompt with explicit shape control. Its inside–outside attention mechanism decouples attention between the object region and the background, enhancing structure consistency and reducing semantic drift in the edited areas. In addition to the attention module, the paper \cite{Training-Free_CS} focuses on the residual connections in the denoising network. It finds that the residual connections in specific encoder layers primarily transmit content information, while the corresponding decoder layers transmit style information. Based on this, the proposed method efficiently and flexibly achieves content and style transfer by selectively injecting these features.

Region-mask-based methods use manually or automatically generated masks to guide localized edits. For instance, Blended Latent Diffusion \cite{Blended_diffusion,Blended_latent_diffusion} requires user-defined masks and employs a progressive mask-shrinkage strategy for smooth transitions. In contrast, other methods focus on automatic mask generation. DiffEdit \cite{Diffedit} computes a mask from the difference in noise predictions between different text prompts. Zone \cite{Zone} creates an instruction-guided coarse mask and refines it using the Segment Anything Model \cite{SAM}. FISEdit \cite{FISEdit} prioritizes efficiency by performing sparse denoising only in semantically altered regions while reusing features elsewhere, enabling fast, high-fidelity edits.

The fundamental limitation of these spatial-domain methods is their heavy reliance on cumbersome empirical observations for the selection of feature maps or attention maps, rather than on theoretical analysis based on clear causal relationships. This reliance on empirical selection results in a lack of theoretical foundation for the correlation between specific spatial feature maps or attention maps and the final image structure, leading to uncertainty and the risk of failure in the intervention outcomes.
\section{Method}
\subsection{Preliminaries}
The proposed method is based on the framework of the Latent Diffusion Model (\emph{i.e.}, LDM). The LDM consists of three core components: a variational autoencoder (\emph{i.e.}, VAE), a text encoder $\bm{\tau}$, and a conditional denoising network $\bm{\epsilon}_\theta$, which serves as the denoiser in the latent space.

The diffusion process takes place in the latent space encoded by the VAE. Given an input image $\bm{x}$, the encoder $\bm{E}$ of the VAE projects it into the latent variable $\bm{z}_0=\bm{E}(\bm{x})$. During the forward diffusion process, $\bm{z}_0$ is progressively corrupted by the noise sampled from the standard Gaussian distribution over $T$ time steps until the distribution of the noisy latent variable $\bm{z}_t$ approximates the standard Gaussian distribution. Leveraging the property of the Gaussian distribution, the noisy latent variable $\bm{z}_t$ at an arbitrary timestep $t$ can be directly sampled based on the initial latent variable $\bm{z}_0$ as follows:
\begin{equation}
\label{eq1}
\bm{z}_t=\sqrt{\bar{\alpha}_t}\bm{z}_0 +\sqrt{1-\bar{\alpha}_t}\bm{\epsilon}\quad\bm{\epsilon}\sim\mathcal{N}(\bm{0},\bm{I}),
\end{equation}
where $\bm{\epsilon}\sim\mathcal{N}(\bm{0},\bm{I})$  represents the noise, and $\bar{\alpha}_t$ serves as a predefined scaling factor that governs the noise ratio schedule.

Conversely, the generation process is the iterative denoising process. The input text prompt $p$ is encoded into the semantic embeddings $\bm{c}=\bm{\tau}(p)$, which are integrated into the denoising network $\bm{\epsilon}_\theta$ by the attention mechanism to guide the generation process. The generation process runs from the noisy latent variable $\bm{z}_T\sim\mathcal{N}(\bm{0}, \bm{I})$. At each time step $t$, the denoising network $\bm{\epsilon}_{\theta}(\bm{z}_t, t,\bm{c})$ predicts the noise of the current noisy latent variable $\bm{z}_t$ to derive the noisy latent variable $\bm{z}_{t-1}$. Following the iterative denoising steps, the final recovered latent variable $\bm{z}_0$ is decoded back into the pixel space image $\bm{x}=\bm{D}(\bm{z}_0)$ by the decoder $\bm{D}$ of the VAE.
\subsection{Frequency Spectrum Analysis of the Diffusion Process}
We investigate the diffusion process from a frequency-domain perspective to uncover the hierarchical emergence mechanism of structure framework and fine-grained texture. Specifically, we conduct a frequency analysis for the forward diffusion and the reverse generative processes and the analysis consists of three progressive steps. (1) characterizing the power spectral distribution of the initial latent representation, (2) characterizing the frequency spectrum degradation dynamics during the forward diffusion, and (3) characterizing the hierarchical emergence of the reverse generative process.

First, extensive studies have confirmed that the power spectral density (\emph{i.e.}, PSD) of natural images follows the power-law distribution \cite{image_psd_1,image_psd_2},
\begin{equation}
\label{eq2}
P(\omega)\propto\omega^{-\beta},
\end{equation}
where $\beta > 0$ and $\omega$ denotes different frequency component. This indicates that the energy of natural images is concentrated in the lower frequency components and decays with increasing frequency. Within the LDM framework, the VAE is designed to balance information compression with reconstruction fidelity. Although the VAE performs nonlinear mappings, its optimization for perceptual consistency necessitates the preservation of essential spatial correlations and structural topology. Consequently, it is reasonable to posit that the latent variable $\bm{z}_0$ effectively inherits the power spectral density distribution of the original image, which means $P_{\bm{z}_0}(\omega)\propto \omega^{-\beta}$.

Second, building upon the power spectral density distribution of $\bm{z}_0$, we quantify how these frequency components of $\bm{z}_0$ degrade during the diffusion process. The noisy latent variable $\bm{z}_t$ constitutes a linear combination of the initial latent variable $\bm{z}_0$ and the Gaussian noise $\bm{\epsilon}$. By leveraging the linearity of the Fourier transform \cite{fft}, the frequency spectrum representation $\mathcal{F}(\bm{z}_t)$ is:
\begin{equation}
\label{eq3}
\mathcal{F}(\bm{z}_t) = \sqrt{\bar{\alpha}_t}\mathcal{F}(\bm{z}_0)+ \sqrt{1-\bar{\alpha}_t}\mathcal{F}(\bm{\epsilon}),
\end{equation}
where $\mathcal{F}$ denotes the Fourier transform.

To quantitatively characterize this degradation, we utilize the frequency-dependent Signal-to-Noise Ratio (\emph{i.e.}, SNR), defined as the ratio of the original signal power spectral to that of the noise. Conceptually, when the SNR corresponding to a specific frequency component $\omega$ falls below a threshold, the information encoded by the frequency component is considered effectively submerged by the noise. Considering the stochastic nature of the noisy latent variable $\bm{z}_t$, we employ the PSD to formalize the energy distribution of these frequency components. Premised on the statistical independence between the initial latent variable $\bm{z}_0$ and the Gaussian noise $\bm{\epsilon}$, combined with the zero-mean property of frequency spectrum for Gaussian noise \cite{dctdiff}, the PSD of the noisy latent variable $\bm{z}_t$ is the weighted superposition of the $P_{\bm{z}_0}$ and the $P_{\bm{\epsilon}}$:
\begin{equation}
\label{eq4}
P_{\bm{z}_t}(\omega) = \bar{\alpha}_tP_{\bm{z}_0}(\omega) + (1-\bar{\alpha}_t)P_{\bm{\epsilon}}(\omega).
\end{equation}

Consequently, the frequency-dependent SNR for $\bm{z}_t$ is:
\begin{equation}
\label{eq5}
\text{SNR}(\omega,t) = \frac{\bar{\alpha}_t P_{\bm{z}_0}(\omega)}{(1-\bar{\alpha}_t) P_{\bm{\epsilon}}(\omega)}.
\end{equation}

Considering that $\bm{\epsilon}$ is Gaussian noise, the $P_{\bm{\epsilon}}(\omega)$ is a frequency-independent constant \cite{random_book}. By integrating the property with the $P_{\bm{z}_0}(\omega)\propto\omega^{-\beta}$, Equation \eqref{eq5} elucidates two governing frequency spectrum dynamics: (1) At any timestep $t$, the $\text{SNR}(\omega,t)$ exhibits a monotonic decrease with respect to frequency component $\omega$, which indicates that higher-frequency components are significantly more susceptible to noise degradation than the lower-frequency components. (2) As the diffusion process progresses, the coefficient $\frac{\bar{\alpha}_t}{1-\bar{\alpha}_t}$ diminishes, which leads to the monotonic decrease of SNR across the entire frequency spectrum.

These characteristics collectively illustrate a frequency-dependent hierarchical degradation process. The higher-frequency spectrum, which contains fine-grained textures, undergoes rapid decay in SNR and is significantly affected by noise during the early diffusion stage. In contrast, the lower-frequency spectrum, responsible for the structural framework, experiences a slower SNR decay and is significantly affected by noise until the later diffusion stage.

Finally, we can deduce the evolution of the generative process, which functions as the inverse of forward diffusion. Since the denoising network $\bm{\epsilon}_\theta$ can effectively retrieve valid information only from frequency components with a sufficiently high SNR, the generation process naturally follows a coarse-to-fine recovery order. In the early generation stage, despite the overall suppression of SNR, the lower-frequency components maintain a relatively higher SNR compared to their higher-frequency counterparts. As a result, the denoising network prioritizes these lower-frequency components to establish the structural framework. On the other hand, as the generation progresses and the SNR improves across the entire frequency spectrum, the denoising network shifts its focus to resolving the higher-frequency components, thereby synthesizing detailed textures on the stabilized structural framework.

\begin{figure}[h!]
\centerline{\includegraphics[width=1.0\linewidth]{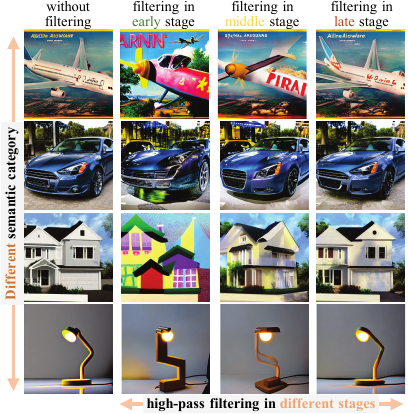}}
\caption{Visual comparison of generation results obtained by applying high-pass filtering to the noisy latent variable $\bm{z}_t$ at different stages. The first column shows the images generated without filtering, whereas the subsequent columns show the images generated when the intervention is applied during the early, middle, and late stages, respectively.}
\label{fig2}
\end{figure}

To empirically substantiate the theoretical deduction, we designed a targeted experiment using Stable Diffusion v1.5 with the 15-step sampling schedule. We segmented the generation process into three distinct stages: early stage (\emph{i.e.}, steps 1--5), middle stage (\emph{i.e.}, steps 6--10), and late stage (\emph{i.e.}, steps 11--15). Within each stage, we apply high-pass filtering separately to the noisy latent variables $\bm{z}_t$ to suppress their low-frequency components, and compare the generated images with those without intervention.

As shown in Fig.~\ref{fig2}, the experimental results support the theoretical insights. Applying high-pass filtering during the early generation stage reveals a significant deviation in the structural framework of the final images compared to those without any intervention. This observation highlights the crucial role that lower-frequency components of the noisy latent variables $\bm{z}_t$ play in establishing the initial structural framework. In contrast, applying high-pass filtering in the later generation stage preserves the structural framework, with variations appearing mainly in the fine-grained textures. These findings emphasize the critical role of the lower-frequency components for establishing the structural framework during the early generation stage, which diminishes as the generation progresses. Concurrently, the denoising network increasingly utilizes higher-frequency components in the later stages to form fine-grained textures upon the established structural framework. Comprehensive experimental details and results are provided in the Appendix.

\begin{figure*}[htp]
\centerline{\includegraphics[width=0.99\linewidth]{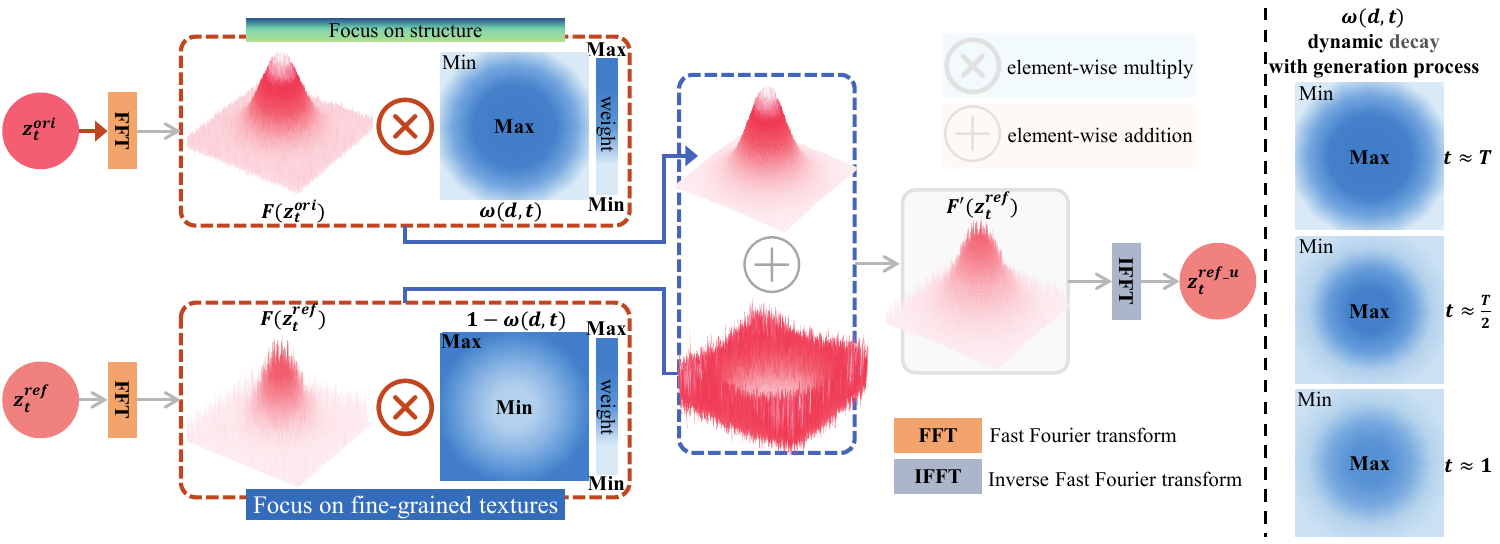}}
\caption{Overview of the frequency modulation method. The method dynamically fuses the frequency components of $\bm{z}_t^{original}$ and $\bm{z}_t^{refined}$ using a frequency-dependent weighting function $\omega(d,t)$ with a dynamic decay strategy. It imposes stronger constraints in the early generation stage to preserve structure while gradually relaxing these constraints in the later generation stage to enhance fine-grained texture synthesis.}
\label{fig3}
\end{figure*}
\subsection{Frequency Modulation Method}
Building upon the theoretical analysis, we introduce the frequency modulation method (\emph{i.e.}, FMM), which directly modulates the frequency components of noisy latent variables instead of using heuristic feature injections like spatial-domain methods. As illustrated in Fig.~\ref{fig3}, the FMM employs a frequency-dependent weighting function $\omega(d,t)$ with a dynamic decay strategy. The method modulates the frequency components of $\bm{z}_t^{ref}$ related to the refined prompt $p_{ref}$ using those of $\bm{z}_t^{ori}$ corresponding to the original prompt $p_{ori}$, ensuring global structure consistency and precise semantic updates.

To perform this modulation, we first transform the noisy latent variables $\bm{z}_t^{ori}$ and $\bm{z}_t^{ref}$ into their frequency representations $\mathcal{F}(\bm{z}_t^{ori})$ and $\mathcal{F}(\bm{z}_t^{ref})$ using the Fourier transform. Next, we combine these frequency representations using the frequency-dependent weighting function $\omega(d,t)$ with a dynamic decay strategy to produce the updated frequency representation $\mathcal{F}'(\bm{z}_t^{ref})$, which can be denoted as: 
\begin{equation}
\label{eq6}
\mathcal{F}'(\bm{z}_t^{ref}) = \omega(d,t) \cdot \mathcal{F}(\bm{z}_t^{ori}) + (1-\omega(d,t)) \cdot \mathcal{F}(\bm{z}_t^{ref}).
\end{equation}

Subsequently, the Inverse Fourier transform maps $\mathcal{F}'(\bm{z}_t^{ref})$ back to the spatial domain, yielding the updated noisy latent variable $\bm{z}_t^{ref\_u}$, which can be represented by:
\begin{equation}
\label{eq7}
\bm{z}_t^{ref\_u} = \mathcal{F}^{-1}(\mathcal{F}'(\bm{z}_t^{ref})).
\end{equation}

The updated noisy latent variable $\bm{z}_t^{ref\_u}$ is fed into the denoising network $\bm{\epsilon}_{\theta}$ for the next step in the generation process, which can be denoted as:
\begin{equation}
\label{eq8}
\bm{z}_{t-1}^{ref} = \bm{\epsilon}_{\theta}(\bm{z}_t^{ref\_u}, \bm{c}_{ref}, t),
\end{equation}
where $\bm{z}_{t-1}^{ref}$ is the noisy latent variable for the next step in the generation process, $\bm{c}_{ref}$ is the text embedding derived from the refined prompt $p_{ref}$ via the text encoder $\bm{\tau}$, and $t$ is the current time step index.

The core of the FMM is the frequency-dependent weighting function $\omega(d,t)$ with a dynamic decay strategy, as defined in Equation~\eqref{eq9}. The function assigns relative weights based on the timestep $t$ and the radial distance $d$ from the center of the frequency spectrum. It is primarily governed by two parameters $\alpha$ and $\sigma$. The parameter $\alpha$ serves as a scaling factor for the radial frequency distance. A larger $\alpha$ results in a flatter weight decay distribution, which allows the frequency spectrum constraint from $\bm{z}_t^{ori}$ to extend over a wider frequency range. Conversely, $\sigma$ serves as the standard deviation of the weighting kernel. A smaller $\sigma$ leads to a sharper weight decay distribution, which restricts the constraint from $\bm{z}_t^{ori}$ to a narrower low-frequency band and results in a weaker influence on the higher-frequency components.
\begin{equation}
\label{eq9}
\omega(d,t)=f_{decay}(t) \cdot \exp((\frac{-d}{max(d)\cdot\alpha})^2\cdot(2\sigma^2)^{-1}).
\end{equation}
Further, we introduce a time-dependent decay factor $f_{decay}(t)$, as defined in Equation~\eqref{eq10}, to regulate the relative weights in line with the evolution of the generation process. During the early generation stage, $f_{decay}(t)$ enhances the weights across the entire frequency spectrum, leading to the frequency components of $\bm{z}_t^{ref}$ being more influenced by those of the $\bm{z}_t^{ori}$. This early enhancement ensures consistency in the structural framework. As the generation process progresses, the frequency constraint on $\bm{z}_t^{ref}$ must be gradually relaxed to accommodate the semantic adjustments specified by the refined prompt $p_{ref}$. Consequently, $f_{decay}(t)$ decreases smoothly. This dynamic decay mechanism allows for the modification of semantics and textures while simultaneously preserving the structural framework.
\begin{equation}
\label{eq10}
f_{decay}(t) = \exp((t-T)\cdot T^{-1}).
\end{equation}

\begin{figure*}[h!]
\centerline{\includegraphics[width=0.9\linewidth]{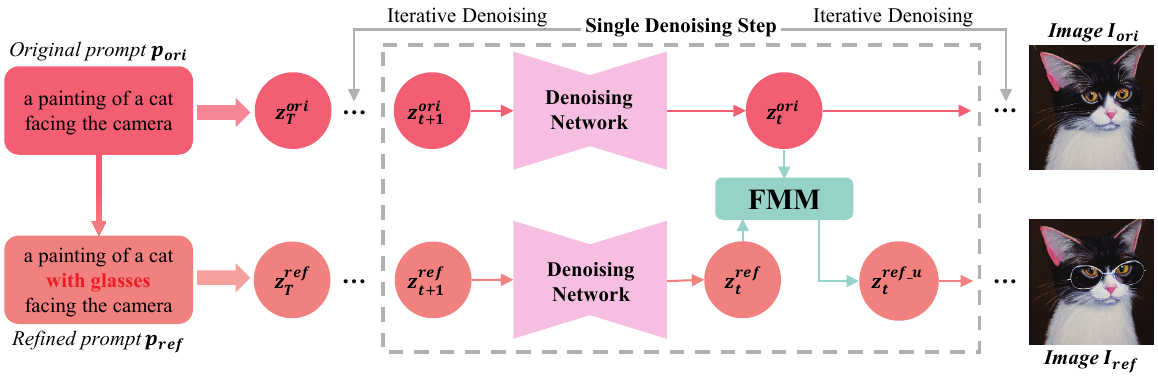}}
\caption{Overall generation pipeline. At each generation step, the FMM dynamically modulates the frequency components of the $z_{t}^{ref}$ using the proposed frequency-dependent weighting function $\omega(d,t)$ with a dynamic decay. Consequently, the final image $I_{ref}$ inherits the structure framework of image $I_{ori}$ while faithfully reflecting the semantic content specified by the text prompt $p_{ref}$.}
\label{fig4}
\end{figure*}

The complete generation pipeline is illustrated in Fig.~\ref{fig4}. Specifically, at each generation step, the FMM operates on the noisy latent variable $z_{t}^{ori}$ and the noisy latent variable $z_{t}^{ref}$. By dynamically fusing the frequency components of the $z_{t}^{ori}$ and the $z_{t}^{ref}$ to modulate the frequency components of the $z_{t}^{ref}$, ensuring that the generated image $I_{ref}$ inherits the structure framework from the image $I_{ori}$ while faithfully manifesting the intended semantic modifications specified by the text prompt $p_{ref}$.

\section{Experiments}
\subsection{Experiments Setup}
\subsubsection{Datasets}
We evaluate the proposed method on two benchmarks: the PIE-Bench \cite{PIE1} and the  ImageNetR-Fake \cite{pnp}. The PIE-Bench consists of 700 paired text prompts with various types of text prompt changes, including local replacement, attribute modification, object addition, \emph{etc}. The ImageNetR-Fake consists of 150 paired text prompts with various types of text prompt changes, including background adjustment, style adjustment, material adjustment, \emph{etc}.
\subsubsection{Comparison Methods}
We use PnP \cite{pnp}, FPE \cite{FPE}, TtfDf \cite{ttfdiffusion}, Diffedit \cite{Diffedit}, and h-edit \cite{h-edit} as comparison methods. PnP and FPE are representative attention-based methods to guide the generation corresponding to the refined prompt. TtfDf employs gradient-based optimization to discover interpretable directions within the bottleneck feature space for semantic adjustment. Diffedit is a mask-based method that derives a semantic mask by comparing noise predictions under different text prompts and applies this mask to confine the updating region. h-edit leverages Doob’s h-transform to disentangle sampling updates into reconstruction and editing terms. For a comprehensive and fair comparison, all methods are evaluated based on the Stable Diffusion v1.5 model.
\subsubsection{Evaluation Metrics}
The generated results are quantitatively evaluated from three complementary dimensions: structure preservation, background consistency, and image-text alignment.

First, the structure preservation is quantified using the structure distance metric \cite{dino_self}. The metric measures the magnitude of structural change between the original and refined images while remaining invariant to appearance variations. A lower structure distance score signifies superior preservation of structural integrity. Second, the background consistency is quantified using the LPIPS \cite{lpips}, PSNR \cite{PSNR}, and SSIM \cite{SSIM}. The lower LPIPS values and higher PSNR and SSIM values indicate higher fidelity in background consistency. Finally, the image-text alignment is measured by the CLIP Score. The metric evaluates the semantic correspondence between the generated refined image and the refined text prompt by computing their feature similarity in the CLIP model embedding space. A higher CLIP Score indicates a more precise alignment with the intended textual semantics.

\subsubsection{Parameter Settings}
We use the deterministic DDIM \cite{DDIM} sampling strategy with 50 generation steps. The classifier-free guidance scale \cite{Classifier-Free} is set to 7.5. The $\alpha$ parameter of the dynamic Gaussian weighting function is set to 0.2, and the $\sigma$ parameter of the dynamic Gaussian weighting function is set to 0.4.

\subsection{Comparison with Baselines}
\subsubsection{Visualization Comparison}
\begin{figure}[htp]
\centerline{\includegraphics[width=1.0\linewidth]{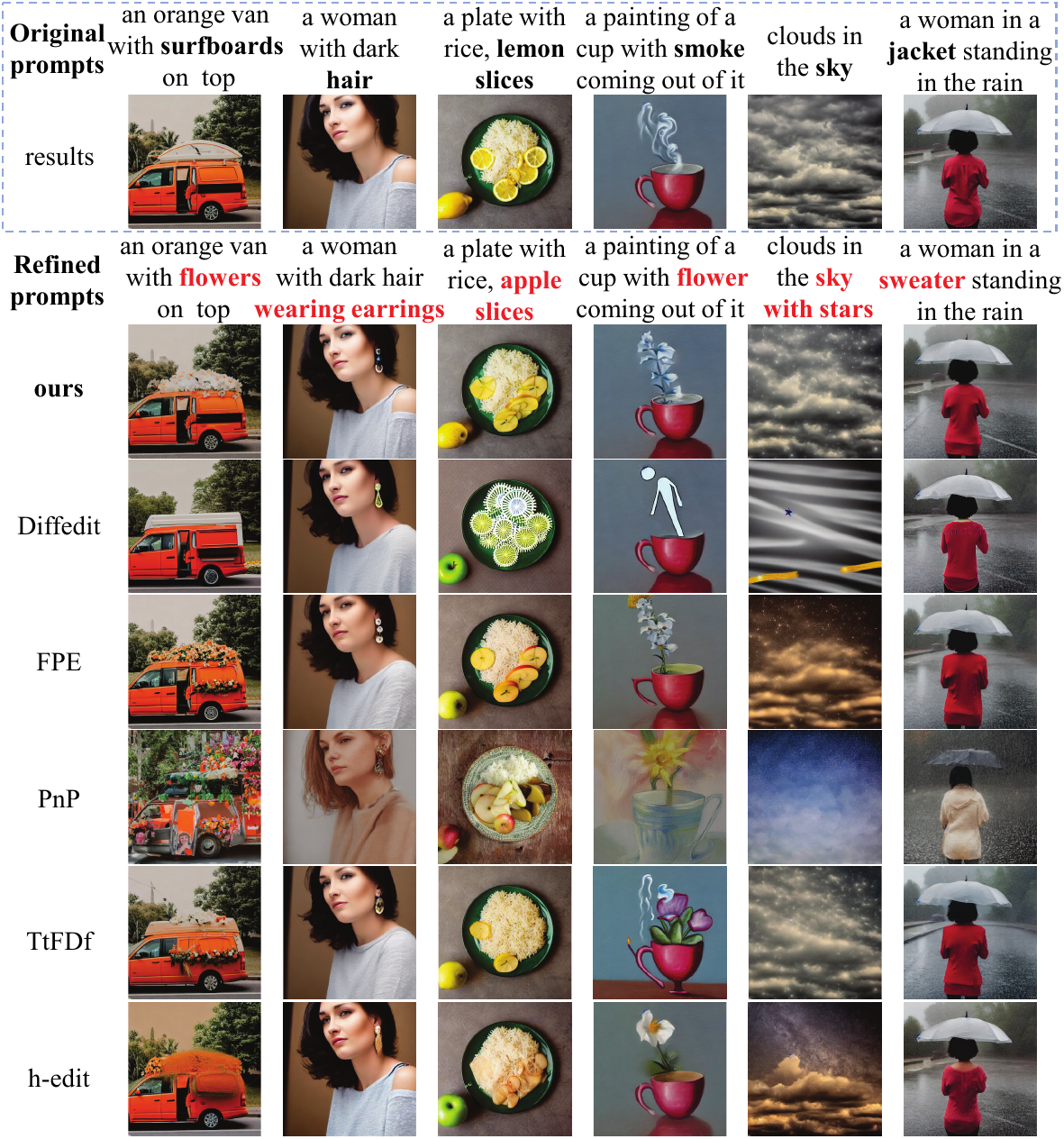}}
\caption{Qualitative comparison on various iterative prompt refinement scenarios. The proposed method well generates the semantic contents specified in the refined prompts while strictly preserving the structure of the original images. In contrast, the other methods either fail to preserve the structure or struggle to render the semantic contents accurately.}
\label{fig5}
\end{figure}

\begin{figure}[htp]
\centerline{\includegraphics[width=1.0\linewidth]{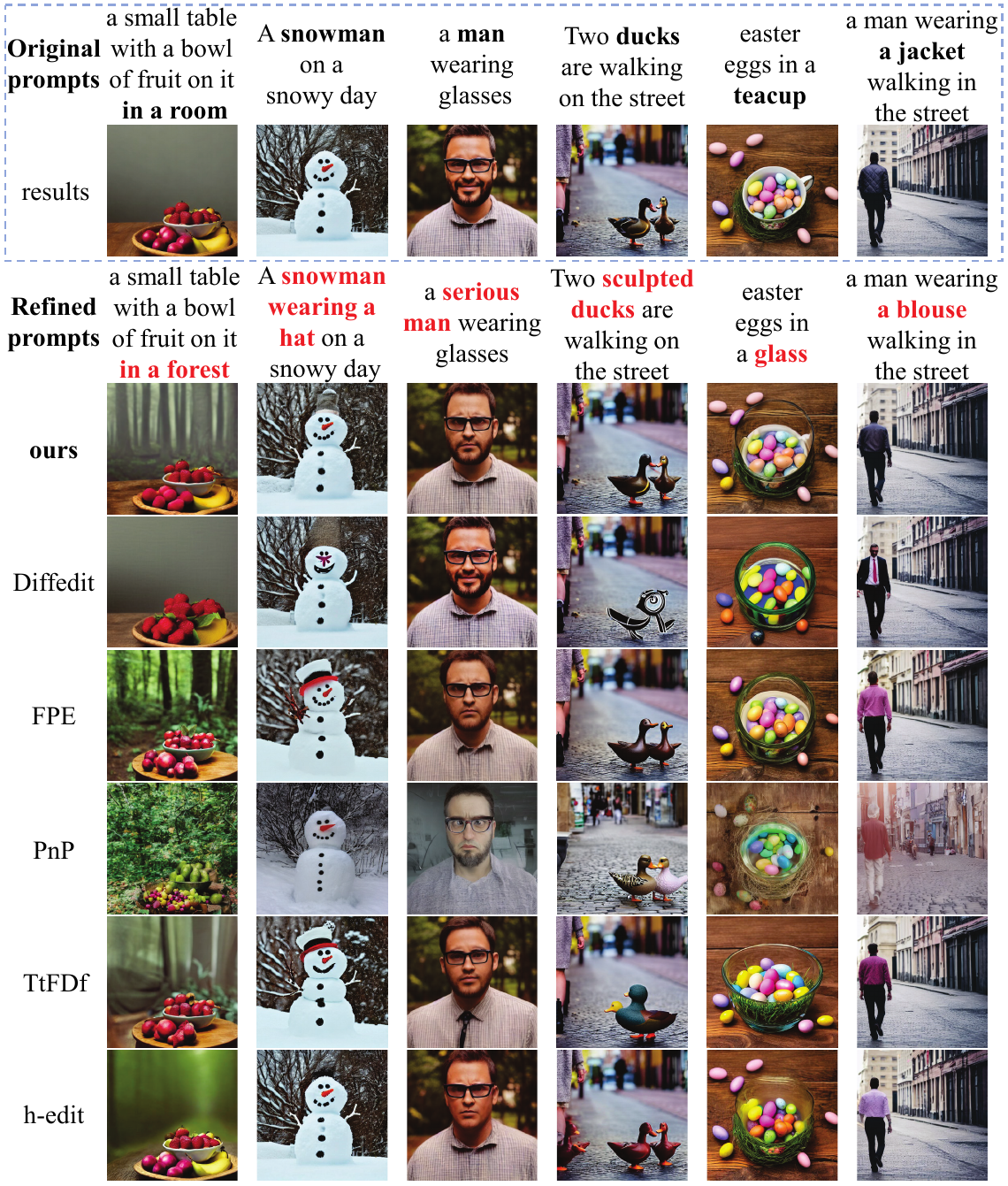}}
\caption{Further qualitative comparisons demonstrating the superiority of the proposed method. The proposed method accurately executes semantic modifications while maintaining high structure fidelity. The other methods often exhibit significant structural degradation or semantic mismatch in these scenarios.}
\label{fig6}
\end{figure}

Fig.~\ref{fig5} and Fig.~\ref{fig6} present the qualitative comparisons across diverse iterative text prompt refinement scenarios. Regardless of the specific adjustments in the text prompts, the proposed method consistently demonstrates robust performance. The generated images exhibit faithful alignment with the refined prompts, synthesizing corresponding semantic content while rigorously preserving the global structure of the initial generation. This capability effectively fulfills the user's creative objectives within the iterative prompt refinement paradigm.

In contrast, the comparison methods exhibit varying degrees of limitations. While the FPE method achieves good semantic alignment with the refined prompts, it struggles to maintain strict consistency in the global structure, often resulting in subtle structural discrepancies between the two images. The DiffEdit method is severely limited by its automatically inferred semantic masks. It produces poor results in terms of semantic alignment and structure preservation when the mask corresponding to the refined prompts cannot be accurately derived. The PnP method relies on the intervention of specified attention maps within the denoising network. When the attention maps are not selected properly, the expected semantic content cannot be generated, and the original global structure cannot be preserved. Although the TtfDf method can generate semantic contents that match the refined prompts to some extent, they are significantly inferior to the proposed method in terms of visual details and preserving the global structure. While the h-edit method demonstrates robust capabilities in maintaining the structure framework, it falls short of the proposed method in terms of background consistency and image-text alignment.

Overall, existing spatial-domain methods struggle to reconcile the conflicting objectives of structure preservation and semantic alignment. In contrast, the proposed Frequency Modulation method offers a more principled solution by addressing structure and semantic content in their respective frequency spectrum distribution, which provides a robust tool for image generation paradigm driven by the iterative refinement of text prompts.

\subsubsection{Quantitative Comparison}
As presented in Table \ref{tab1}, our method achieves excellent performance in structure preservation and background consistency, achieving the optimal or suboptimal results in the metrics. Concurrently, the proposed method also demonstrates highly competitive performance in image-text alignment.

\begin{table}[]
\centering
\caption{The quantitative evaluation results. The optimal results are marked in red, and the suboptimal results are marked in blue. Str. D. means the structure distance and CLIP S. means the CLIP Score.}
\label{tab1}
\begin{tabular}{@{}ccccccc@{}}
\toprule
Dataset                           & Method        & Str. D.($\downarrow$)                   & CLIP S.($\uparrow$)                     & LPIPS($\downarrow$)                    & PSNR($\uparrow$)                        & SSIM($\uparrow$)                       \\ \midrule
& TtfDf         & 42.4081                                 & 26.0262                                 & 0.2240                                 & 17.3285                                 & 0.6499                                 \\
& FPE           & 35.0359                                 & {\color[HTML]{FE0000} \textbf{26.7253}} & {\color[HTML]{3166FF} 0.1849}          & {\color[HTML]{3166FF} 18.7590}          & 0.7042                                 \\
& h-edit        & {\color[HTML]{FE0000} \textbf{32.8434}} & 26.3929                                 & 0.1919                                 & 17.6958                                 & {\color[HTML]{3166FF} 0.7174}          \\
& PnP           & 45.1111                                 & 26.4849                                 & 0.4742                                 & {\color[HTML]{333333} 12.6653}          & 0.3314                                 \\
& Diffedit      & 46.1566                                 & 21.6037                                 & 0.2945                                 & 15.8990                                 & 0.5559                                 \\
\multirow{-6}{*}{\rotatebox[origin=c]{90}{PIE-Bench}}       & \textbf{ours} & {\color[HTML]{3166FF} 33.5045}          & {\color[HTML]{3166FF} 26.6235}          & {\color[HTML]{FE0000} \textbf{0.1425}} & {\color[HTML]{FE0000} \textbf{20.8003}} & {\color[HTML]{FE0000} \textbf{0.7409}} \\ \midrule
& TtfDf         & 49.6320                                 & 25.6117                                 & 0.3471                                 & 13.5393                                 & 0.4924                                 \\
& FPE           & {\color[HTML]{3166FF} 44.6252}          & {\color[HTML]{FE0000} \textbf{27.1672}} & {\color[HTML]{3166FF} 0.3158}          & {\color[HTML]{3166FF} 14.5546}          & 0.5140                                 \\
& h-edit        & {\color[HTML]{333333} 44.7574}          & 26.3490                                 & 0.3190                                 & 13.8001                                 & {\color[HTML]{FE0000} \textbf{0.5609}} \\
& PnP           & 47.7441                                 & 26.6620                                 & 0.5612                                 & 10.6050                                 & 0.2809                                 \\
& Diffedit      & 49.6192                                 & 21.1502                                 & 0.3486                                 & 13.0628                                 & 0.4753                                 \\
\multirow{-6}{*}{\rotatebox[origin=c]{90}{ImageNetR-Fake}} & \textbf{ours} & {\color[HTML]{FE0000} \textbf{44.3462}} & {\color[HTML]{3166FF} 26.9430}          & {\color[HTML]{FE0000} \textbf{0.2712}} & {\color[HTML]{FE0000} \textbf{15.8880}} & {\color[HTML]{3166FF} 0.5328}          \\ \bottomrule
\end{tabular}
\end{table}

In contrast, the comparison methods all exhibit corresponding limitations. Although the FPE achieves a slightly higher CLIP Score on both datasets, its performance in both structure preservation and background consistency is weaker, indicating that its generated results still have detailed discrepancies in composition preservation compared to the original images. While the CLIP Score of the PnP is close to that of the proposed method, its structure preservation and background consistency are significantly weaker, which shows a substantial difference in composition preservation between its generated results and the original images. The Diffedit performs the worst across all metrics, with a particularly low CLIP Score, suggesting that its reliance on the auto mask generation mechanism is highly unstable and often fails to fulfill user intent in diverse iterative refinement scenarios. Although the structure preservation metric of the h-edit is slightly better than that of our method on the PIE dataset, its CLIP Score metric and background consistency metrics are still inferior to those of our method.

These experimental results demonstrate that the proposed frequency modulation method achieves the optimal balance between preserving global composition and realizing precise semantic updates, and its overall performance is superior to current spatial-domain-based comparison methods.

\section{Discussion}
\subsection{Impact of Parameter $\alpha$}
\begin{figure}[htp]
\centerline{\includegraphics[width=1.0\linewidth]{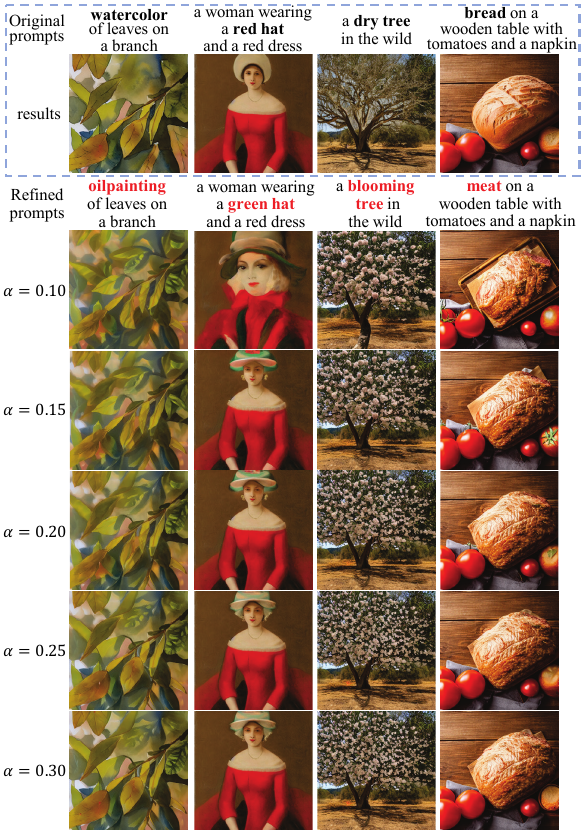}}
\caption{Visualization comparison of the generation results using different $\alpha$ parameter values}
\label{fig7}
\end{figure}

To investigate the influence of $\alpha$, we fixed $\sigma$ at 0.40 and varied $\alpha$ within the range of $[0.10, 0.30]$. The experiment is conducted on the PIE dataset. As presented in Table \ref{tab2} and Fig.~\ref{fig7}, the variation of $\alpha$ reveals the trade-off between structure and background preservation and semantic content consistency in the generated results. When $\alpha$ increases from 0.10 to 0.30, structural preservation and background consistency show an improvement trend. In contrast, the CLIP Score, which represents text-image alignment, consistently decreases.

In the frequency-dependent weighting function $\omega(d,t)$ with dynamic decay, the parameter $\alpha$ acts as a scaling factor for the relative value of frequency radial distance. A larger $\alpha$ value results in smaller relative frequency radial distances $\frac{d}{d_{max}}$ for all frequency components in the exponential term, thereby increasing the weight allocated to all frequency components in the frequency domain. This leads to a flatter weight distribution of the frequency-dependent weighting function across the entire frequency domain. This means that the original noisy latent variable $\bm{z}_t^{ori}$ can impose constraints on the iteratively modified latent variable $\bm{z}_t^{ref}$ across a broader frequency range. Such a stronger frequency-domain constraint naturally retains more of the global composition and background information of the original generated image. However, this also limits the ability to synthesize high-frequency semantic details based on the newly optimized text prompt during re-generation, leading to a decrease in text-image alignment.

\begin{table}[]
\centering
\caption{The influence of different parameter $\alpha$ on the generated results. The parameter $\sigma$ is fixed at 0.4.}
\label{tab2}
\setlength{\tabcolsep}{0.3mm}{\begin{tabular}{@{}ccccccc@{}}
\toprule
             & Str. D.($\downarrow$) & CLIP S.($\uparrow$) & LPIPS($\downarrow$) & PSNR($\uparrow$) & SSIM($\uparrow$)  \\ \midrule
$\alpha=0.10$ & 41.0797                          & 27.1349                & 0.2067              & 18.1732          & 0.5328  \\
$\alpha=0.15$ & 36.6288                          & 26.8279                & 0.1681              & 19.6578          & 0.4924  \\
$\alpha=0.20$ & 33.5045                          & 26.6235                & 0.1425              & 20.8003          & 0.5140  \\
$\alpha=0.25$ & 30.7878                          & 26.4786                & 0.1215              & 21.8049          & 0.2809  \\
$\alpha=0.30$ & 28.3551                          & 26.3281                & 0.1048              & 22.6794          & 0.4040  \\ \bottomrule
\end{tabular}}
\end{table}

\subsection{Impact of Parameter $\sigma$}
\begin{figure}[htp]
\centerline{\includegraphics[width=1.0\linewidth]{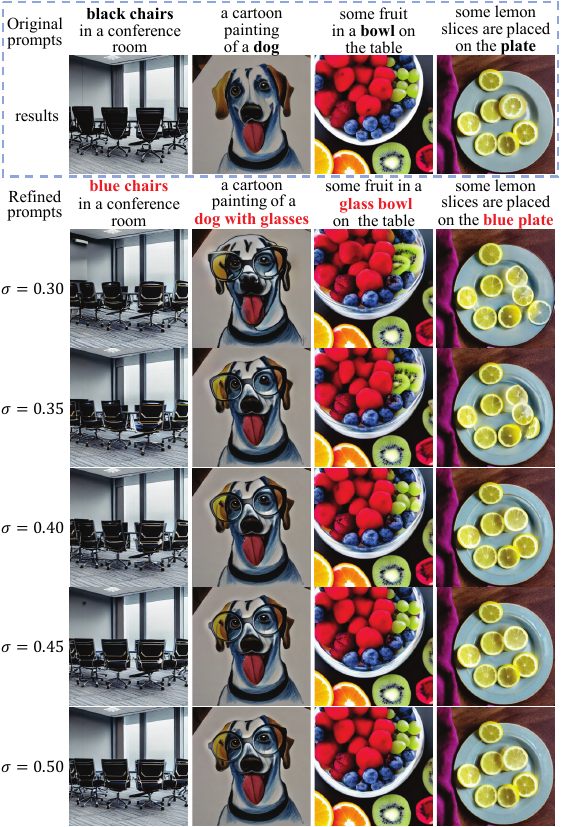}}
\caption{Visualization comparison of the generation results using different $\sigma$ parameter values}
\label{fig8}
\end{figure}

To investigate the influence of parameter $\sigma$, we fixed $\alpha$ at 0.20 while varying $\sigma$ within the range of $[0.30, 0.50]$. The experiment is conducted on the PIE dataset. As illustrated in Table \ref{tab3} and Fig.~\ref{fig8}, the variation of $\sigma$ similarly reveals the trade-off between structural and background preservation and semantic content consistency in the generated results. When $\sigma$ increases from 0.30 to 0.50, structural preservation and background consistency show an improvement trend. In contrast, the CLIP Score, which represents text-image alignment, consistently decreases.

In the frequency-dependent weighting function $\omega(d,t)$ with dynamic decay, the parameter $\sigma$ acts as the standard deviation of the weighting kernel, influencing the concentration of the weight distribution over the frequency components. A smaller $\sigma$ value causes the frequency-dependent weighting function to allocate smaller weights to all frequency components, leading to a more severe decay in the weight distribution across the entire frequency domain. The weight becomes more concentrated on the low-frequency components at the center of the frequency spectrum, with the weights of the mid- and high-frequency components decaying rapidly. This means that the constraint imposed by the original noisy latent variable $\bm{z}_t^{ori}$ on the iteratively modified latent variable $\bm{z}_t^{ref}$ is more limited to the low-frequency.

\begin{table}[]
\centering
\caption{The influence of different parameter $\sigma$ on the generated results. The parameter $\alpha$ is fixed at 0.2.}
\label{tab3}
\setlength{\tabcolsep}{0.3mm}{\begin{tabular}{@{}ccccccc@{}}
\toprule
             & Str. D.($\downarrow$) & CLIP S.($\uparrow$) & LPIPS($\downarrow$) & PSNR($\uparrow$) & SSIM($\uparrow$)  \\ \midrule
$\sigma=0.30$ & 36.6213                          & 26.8281                & 0.1681              & 19.6582          & 0.7060   \\
$\sigma=0.35$ & 34.8020                          & 26.7450                & 0.1533              & 20.2762          & 0.7255   \\
$\sigma=0.40$ & 33.5045                          & 26.6235                & 0.1425              & 20.8003          & 0.7409   \\
$\sigma=0.45$ & 32.1494                          & 26.5880                & 0.1322              & 21.2877          & 0.7563   \\
$\sigma=0.50$ & 30.7899                          & 26.4820                & 0.1215              & 21.8067          & 0.7724   \\ \bottomrule
\end{tabular}}
\end{table}

\subsection{Impact of Weighting Function}
The core of the proposed FMM is the frequency-dependent weighting function $\omega(d,t)$. This function dynamically assigns relative weights to the frequency components based on the generation timestep $t$ and the radial distance $d$ from the center of the frequency spectrum. We formulate the weighting function as a Gaussian distribution form, as defined in Equation~\eqref{eq9}. We specifically opted for this Gaussian distribution form due to its smooth, asymptotic decay properties, which ensure a continuous transition of the frequency spectrum without introducing abrupt discontinuities. To validate the effectiveness of this design, we conducted an ablation study on the PIE dataset by comparing it against a linear weighting function. In contrast to the smooth and asymptotic decay of the Gaussian weighting function, the linear weighting function imposes a constant linear decline characterized by a hard frequency cutoff, formally defined as Equation~\eqref{eq11}. To ensure a rigorous comparison, we set the parameter $\alpha$ to 0.2 for the linear weighting function, which is the same as the parameter $\alpha$ value used by our proposed Gaussian weighting function.

\begin{equation}
\label{eq11}
\omega(d, t)=f_{decay}(t)\cdot\max\left(0, 1-\frac{d}{max(d)\cdot\alpha}\right).
\end{equation}

\begin{figure}[htp]
	\centerline{\includegraphics[width=1\linewidth]{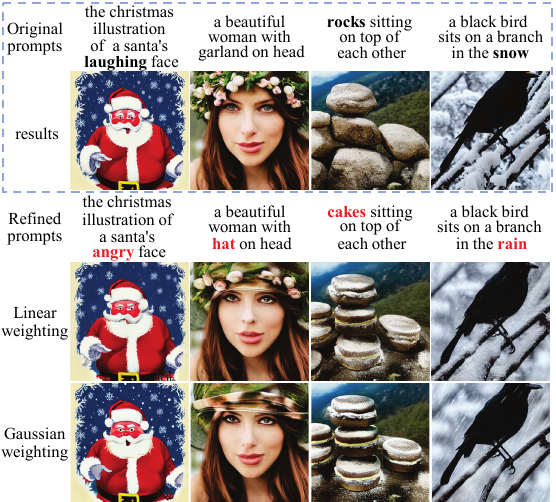}}
	\caption{Visual comparison of the generation results using the Gaussian weighting function and the linear weighting function.}
	\label{fig9}
\end{figure}

As illustrated in Table \ref{tab4} and Fig.~\ref{fig9}, while the frequency modulation based on the linear weighting function yields comparable performance in terms of structure preservation and background consistency, a significant performance gap appears in image-text alignment. The frequency modulation based on the linear weighting function achieves a significantly lower CLIP Score compared to Gaussian weighting function. These results reveal the fundamental limitation of the linear weighting function.

\begin{table}[]
\centering
\caption{The influence of different weighting function forms  on the generated results. We compare the linear weighting function with our proposed Gaussian weighting function under the same $\alpha$ parameter value 0.2.}
\label{tab4}
\setlength{\tabcolsep}{0.3mm}{\begin{tabular}{@{}cccccc@{}}
\toprule
			 & Str. D.($\downarrow$) & CLIP S.($\uparrow$) & LPIPS($\downarrow$) & PSNR($\uparrow$) & SSIM($\uparrow$)  \\ \midrule
Linear    & 34.1746                          & 25.8356                & 0.1496              & 20.6710          & 0.7252  \\
\textbf{Gaussian} & {\color[HTML]{000000} \textbf{33.5045}} & {\color[HTML]{000000} \textbf{26.6235}} & {\color[HTML]{000000} \textbf{0.1425}} & {\color[HTML]{000000} \textbf{20.8003}} & {\color[HTML]{000000} 0.7409} \\ \bottomrule
\end{tabular}}
\end{table}

First, although the linear weighting function imposes sufficient constraints on low-frequency components to preserve global structure, its uniform decay pattern lacks spectrum selectivity. The uniform decay fails to optimally modulate different spectrum components to effectively distinguish the coarse structure framework and fine-grained semantic content, thereby hindering the synthesis of the semantic content required by the refined prompt.

Second, the linear weighting function characterized by a hard frequency cutoff introduces discontinuities within the frequency domain. The cutoff disturbs the coherence between the preserved structure and the newly synthesized semantic content. In contrast, the smooth, asymptotic nature of the Gaussian function ensures a continuous spectrum modulation, which enables the natural integration of structure constraints with newly synthesized semantic content.

In conclusion, while frequency modulation based on a linear weighting function yields comparable performance in structure preservation and background consistency, its lack of spectral selectivity and the discontinuities induced by the hard frequency cutoff significantly hinder the intended semantic updates. In contrast, our proposed Gaussian-based frequency modulation proves to be the superior design. By offering a smooth, continuous, and inherently selective spectral modulation, it facilitates a better balance between structural preservation and the synthesis of new semantic content.
\subsection{Potential and Limitations in Real Image Editing}
\begin{figure*}[htp]
	\centerline{\includegraphics[width=0.98\linewidth]{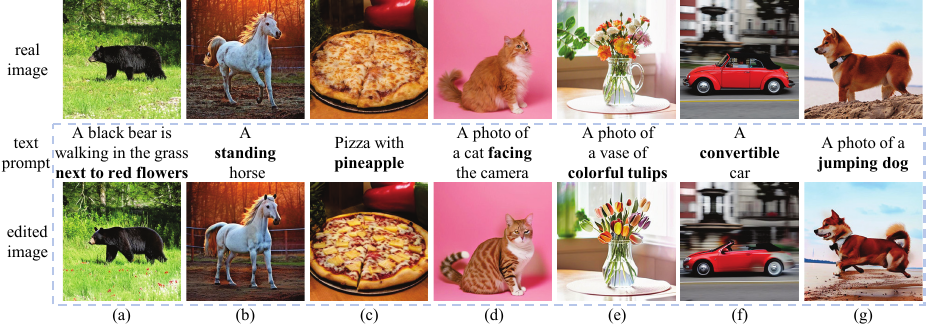}}
	\caption{Visual results of the proposed method applied to the real image editing task. The method demonstrates significant generalization potential, successfully performing attribute modifications (e.g., (a) and (c)) and pose alterations (e.g., (b)) while preserving global structure. However, limitations caused by inversion perturbations are also observed, manifesting as texture discrepancies (d), detail blurring and structural artifacts ((e), (f)), and background inconsistency (g).}
	\label{fig10}
\end{figure*}

To further evaluate our method, we extend it to real image editing using DDIM Inversion to obtain initial noisy latent. As shown in Fig.~\ref{fig10}, our method demonstrates generalization in both attribute modification (e.g., Fig.~\ref{fig10}(a,c)) and pose alteration (e.g., Fig.~\ref{fig10}(b,d)), consistently preserving structure and background. However, limitations persist in the form of texture discrepancy, detail blurring, and structure deviation. Fig.~\ref{fig10}(d) exhibits mismatched fur texture despite preserved structure, while Fig.~\ref{fig10}(e-g) show missing details (e.g., vase handle, driver) and background distortions.

These failures primarily stem from the inexact approximation of the diffusion inversion process, which introduces random and non-uniform frequency spectrum perturbation. Since our method relies on modulating the frequency components of the new noisy latent by fusing it with frequency spectrum information from the original noisy latent, these perturbations interfere with the modulation. Specifically, when inversion noise corrupts high-frequency components, the modulation process introduces artifacts, leading to texture discrepancies or blurring. Conversely, corruption in low-frequency components results in inaccurate structural guidance, leading to structural deviations. Future work could address this by developing inversion-perturbation-aware adaptive frequency modulation, or by employing learning-based methods to repair the non-uniform spectral perturbations in the inverted latents.

\section{Conclusion}
This paper introduces a novel perspective on the generation mechanisms within text-guided diffusion models by focusing on the frequency domain. It highlights how the frequency spectrum of noisy latent variables significantly influences the hierarchical development of image structure and fine-grained textures. In the early generation stage, lower-frequency components play a crucial role in establishing the structural framework. As the generation progresses, their significance wanes, giving way to higher-frequency components that drive the generation of fine-grained textures. Building on this understanding, we proposes the Frequency Modulation Method. The method utilizes a frequency-dependent weighting function with dynamic decay to maintain structure consistency across the frequency spectrum. Furthermore, it incorporates a temporally adaptive relaxation strategy to ensure the evolution of semantic content aligns with the refined prompts. By working directly on the noisy latent variables, this method avoids the fragility associated with choosing internal feature maps for spatial-domain methods. Extensive experiments demonstrate that the method significantly outperforms the state-of-the-art methods, achieving a superior balance between global structure preservation and precise semantic adjustments. This advancement offers valuable insights for enhancing controllability in image generation through iterative prompt refinement. Additionally, the extension to real image editing task reveals both its potential and limitations, suggesting that future research could focus on developing inversion-perturbation-aware adaptive frequency modulation strategies to enhance robustness in practical applications.

\begin{figure*}[h!]
\centerline{\includegraphics[width=0.98\linewidth]{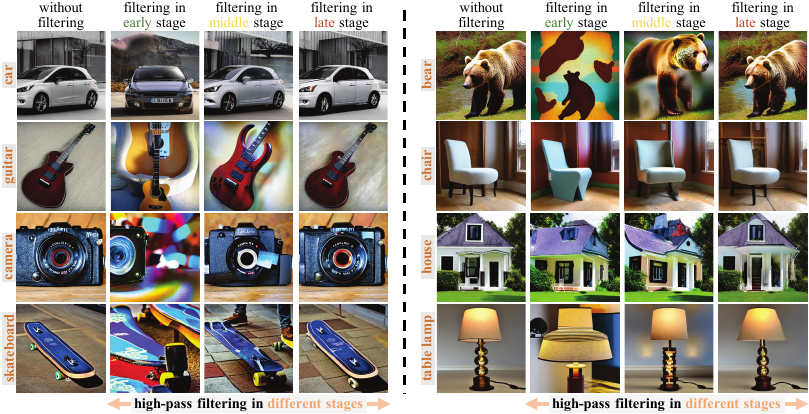}}
\caption{Comparison of the normally generated images with the images generated by performing high-pass filtering at different stages.}
\label{fig11}
\end{figure*}

\begin{figure}[h!]
\centerline{\includegraphics[width=0.99\linewidth]{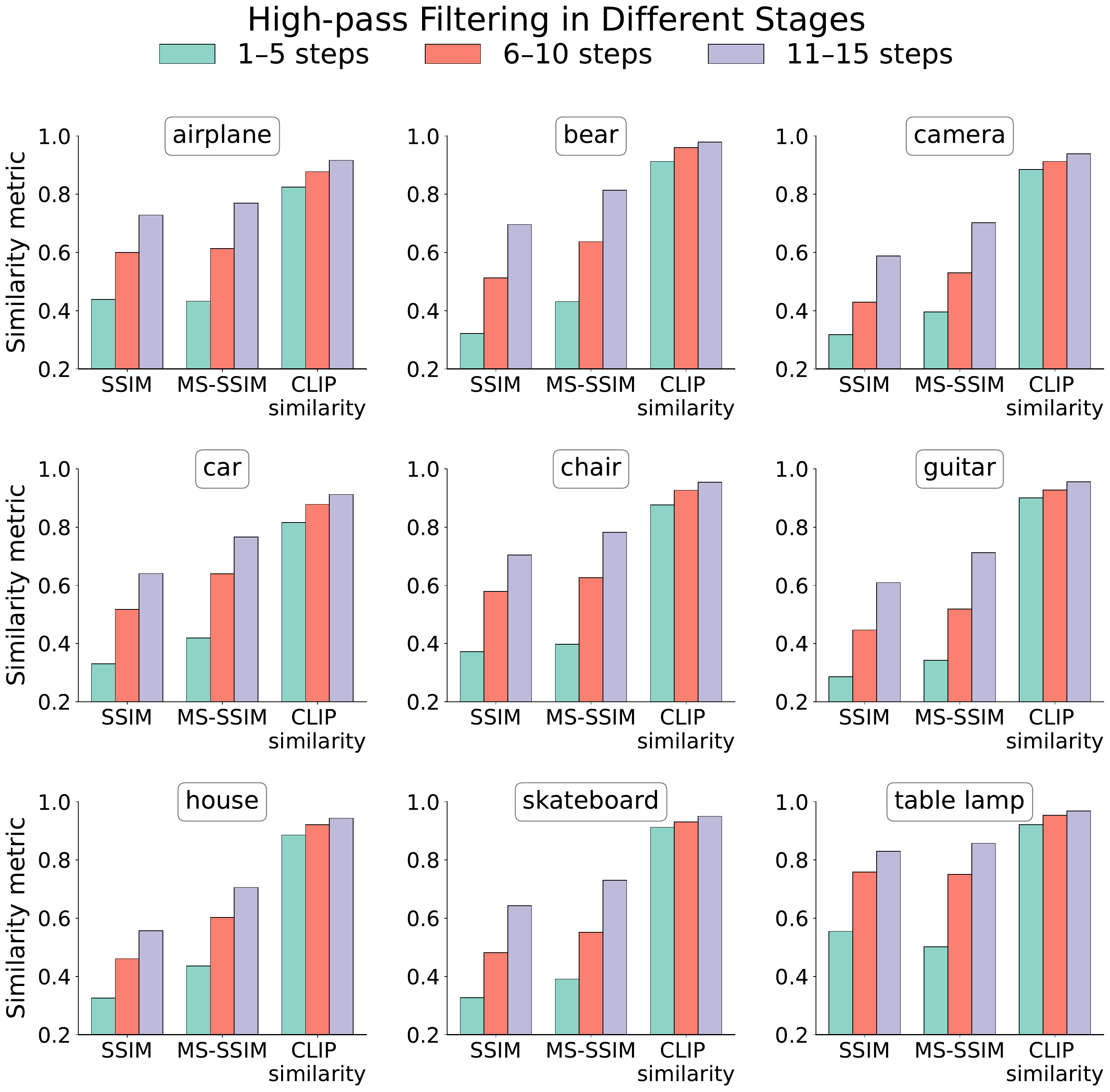}}
\caption{The images generated by performing high-pass filtering at the early stage (i.e., steps 1-5) exhibit the lowest similarity to the corresponding reference images. As high-pass filtering is performed at later stages (i.e. steps 6-10 or 11-15), the similarity gradually increases. These results confirm our key observation.}
\label{fig12}
\end{figure}
\section{Append}
\subsection{Experiment Setup}
We adopted the Stable Diffusion v1.5 model and set the generation steps to 15 with DDIM sampler. We use the words ”airplane”, ”bear”, ”camera”, ”car”, ”chair”, ”guitar”, ”house”, ”skateboard”, ”table lamp” to construct the text prompt ”a  photo of a [x]”. For each text prompt, we set 1000 different random seeds and generate four sets of images. Among them, the first group is normal generation, that is, no high-pass filtering is applied, as the reference images. The second, third, and fourth groups perform high-pass filtering on the noisy latent variables $\bm{z}_t$ at steps 1-5, 6-10, and 11-15 of the generation process, respectively. After that, we compute the similarity between the images generated after high-pass filtering and the normally generated images.
Specifically, we used the SSIM, MS-SSIM \cite{MS-SSIM} and CLIP similarity as metric. The values are between 0 and 1, where a higher value indicates a higher similarity between the two images.
\subsection{Results and Analysis}
Figure.~\ref{fig11} shows the results. The first column shows the normally generated images,  The following three columns show the images generated by performing high-pass filtering in the early stage (i.e., steps 1-5), middle stage (i.e., steps 6-10), and late stage (i.e., steps 11-15) in the generation process, respectively, while keeping the initial noise and text prompts unchanged.

The quantitative results are presented in Figure.~\ref{fig12}. First, images generated by performing high-pass filtering in the early stage always show the lowest similarity to their reference images, indicating that removing lower-frequency components early markedly alters the global structure of the final image. Second, the similarity gradually increases when the same filtering is applied at later stages, suggesting that the global structure is largely determined in the early stage and becomes progressively less sensitive to the lower-frequency components as generation proceeds. These experimental results empirically confirm the theoretical analysis and provide the fundamental motivation for the proposed frequency modulation method. 

\bibliography{reference}

\begin{thebibliography}{10}
\providecommand{\url}[1]{#1}
\csname url@samestyle\endcsname
\providecommand{\newblock}{\relax}
\providecommand{\bibinfo}[2]{#2}
\providecommand{\BIBentrySTDinterwordspacing}{\spaceskip=0pt\relax}
\providecommand{\BIBentryALTinterwordstretchfactor}{4}
\providecommand{\BIBentryALTinterwordspacing}{\spaceskip=\fontdimen2\font plus
\BIBentryALTinterwordstretchfactor\fontdimen3\font minus
  \fontdimen4\font\relax}
\providecommand{\BIBforeignlanguage}[2]{{%
\expandafter\ifx\csname l@#1\endcsname\relax
\typeout{** WARNING: IEEEtran.bst: No hyphenation pattern has been}%
\typeout{** loaded for the language `#1'. Using the pattern for}%
\typeout{** the default language instead.}%
\else
\language=\csname l@#1\endcsname
\fi
#2}}
\providecommand{\BIBdecl}{\relax}
\BIBdecl

\bibitem{DiT}
W.~Peebles and S.~Xie, ``Scalable diffusion models with transformers,'' in
  \emph{Proceedings of the {IEEE/CVF} International Conference on Computer
  Vision}, 2023, pp. 4172--4182.

\bibitem{Photorealistic}
C.~Saharia, W.~Chan, S.~Saxena, L.~Li, J.~Whang, E.~L. Denton, K.~Ghasemipour,
  R.~Gontijo~Lopes, B.~Karagol~Ayan, T.~Salimans \emph{et~al.},
  ``Photorealistic text-to-image diffusion models with deep language
  understanding,'' in \emph{Proceedings of the Conference on Neural Information
  Processing Systems}, 2022, pp. 36\,479--36\,494.

\bibitem{tip_assess}
Z.~Wang, B.~Hu, M.~Zhang, J.~Li, L.~Li, M.~Gong, and X.~Gao, ``Diffusion
  model-based visual compensation guidance and visual difference analysis for
  no-reference image quality assessment,'' \emph{IEEE Transactions on Image
  Processing}, vol.~34, pp. 263--278, 2025.

\bibitem{tip_fusion1}
Y.~Shi, Y.~Liu, J.~Cheng, Z.~J. Wang, and X.~Chen, ``Vdmufusion: A versatile
  diffusion model-based unsupervised framework for image fusion,'' \emph{IEEE
  Transactions on Image Processing}, vol.~34, pp. 441--454, 2025.

\bibitem{tip_image_edit}
T.~Xia, Y.~Zhang, T.~Liu, and L.~Zhang, ``Consistent image layout editing with
  diffusion models,'' \emph{IEEE Transactions on Image Processing}, vol.~34,
  pp. 6978--6992, 2025.

\bibitem{GAN}
I.~J. Goodfellow, J.~Pouget{-}Abadie, M.~Mirza, B.~Xu, D.~Warde{-}Farley,
  S.~Ozair, A.~C. Courville, and Y.~Bengio, ``Generative adversarial nets,'' in
  \emph{Proceedings of the Conference on Neural Information Processing
  Systems}, 2014, pp. 2672--2680.

\bibitem{VAE}
D.~P. Kingma and M.~Welling, ``Auto-encoding variational bayes,'' in
  \emph{Proceedings of the International Conference on Learning
  Representations}, 2014.

\bibitem{LDM}
R.~Rombach, A.~Blattmann, D.~Lorenz, P.~Esser, and B.~Ommer, ``High-resolution
  image synthesis with latent diffusion models,'' in \emph{Proceedings of the
  {IEEE/CVF} Conference on Computer Vision and Pattern Recognition}, 2022, pp.
  10\,674--10\,685.

\bibitem{p2p}
A.~Hertz, R.~Mokady, J.~Tenenbaum, K.~Aberman, Y.~Pritch, and D.~Cohen{-}Or,
  ``Prompt-to-prompt image editing with cross-attention control,'' in
  \emph{Proceedings of the International Conference on Learning
  Representations}, 2023.

\bibitem{pnp}
N.~Tumanyan, M.~Geyer, S.~Bagon, and T.~Dekel, ``Plug-and-play diffusion
  features for text-driven image-to-image translation,'' in \emph{Proceedings
  of the {IEEE/CVF} Conference on Computer Vision and Pattern Recognition},
  2023, pp. 1921--1930.

\bibitem{FPE}
B.~Liu, C.~Wang, T.~Cao, K.~Jia, and J.~Huang, ``Towards understanding cross
  and self-attention in stable diffusion for text-guided image editing,'' in
  \emph{Proceedings of the {IEEE/CVF} Conference on Computer Vision and Pattern
  Recognition}, 2024, pp. 7817--7826.

\bibitem{diffusionclip}
G.~Kim, T.~Kwon, and J.~C. Ye, ``Diffusionclip: Text-guided diffusion models
  for robust image manipulation,'' in \emph{Proceedings of the {IEEE/CVF}
  Conference on Computer Vision and Pattern Recognition}, 2022, pp. 2426--2435.

\bibitem{Instructpix2pix}
T.~Brooks, A.~Holynski, and A.~A. Efros, ``Instructpix2pix: Learning to follow
  image editing instructions,'' in \emph{Proceedings of the {IEEE/CVF}
  Conference on Computer Vision and Pattern Recognition}, 2023, pp.
  18\,392--18\,402.

\bibitem{null_text_inversion}
R.~Mokady, A.~Hertz, K.~Aberman, Y.~Pritch, and D.~Cohen-Or, ``Null-text
  inversion for editing real images using guided diffusion models,'' in
  \emph{Proceedings of the {IEEE/CVF} Conference on Computer Vision and Pattern
  Recognition}, 2023, pp. 6038--6047.

\bibitem{clip}
A.~Radford, J.~W. Kim, C.~Hallacy, A.~Ramesh, G.~Goh, S.~Agarwal, G.~Sastry,
  A.~Askell, P.~Mishkin, J.~Clark \emph{et~al.}, ``Learning transferable visual
  models from natural language supervision,'' in \emph{Proceedings of the
  International Conference on Machine Learning}, 2021, pp. 8748--8763.

\bibitem{Dreaminpainter}
S.~Xie, Y.~Zhao, Z.~Xiao, K.~C.~K. Chan, Y.~Li, Y.~Xu, K.~Zhang, and T.~Hou,
  ``Dreaminpainter: Text-guided subject-driven image inpainting with diffusion
  models,'' \emph{CoRR}, vol. abs/2312.03771, 2023.

\bibitem{Smartbrush}
S.~Xie, Z.~Zhang, Z.~Lin, T.~Hinz, and K.~Zhang, ``Smartbrush: Text and shape
  guided object inpainting with diffusion model,'' in \emph{Proceedings of the
  {IEEE/CVF} Conference on Computer Vision and Pattern Recognition}, 2023, pp.
  22\,428--22\,437.

\bibitem{Dreambooth}
N.~Ruiz, Y.~Li, V.~Jampani, Y.~Pritch, M.~Rubinstein, and K.~Aberman,
  ``Dreambooth: Fine tuning text-to-image diffusion models for subject-driven
  generation,'' in \emph{Proceedings of the {IEEE/CVF} Conference on Computer
  Vision and Pattern Recognition}, 2023, pp. 22\,500--22\,510.

\bibitem{An_image_one_world}
R.~Gal, Y.~Alaluf, Y.~Atzmon, O.~Patashnik, A.~H. Bermano, G.~Chechik, and
  D.~Cohen-or, ``An image is worth one word: Personalizing text-to-image
  generation using textual inversion,'' in \emph{Proceedings of the
  International Conference on Learning Representations}, 2023.

\bibitem{attention}
A.~Vaswani, N.~Shazeer, N.~Parmar, J.~Uszkoreit, L.~Jones, A.~N. Gomez,
  {\L}.~Kaiser, and I.~Polosukhin, ``Attention is all you need,'' in
  \emph{Proceedings of the Conference on Neural Information Processing
  Systems}, 2017, pp. 5998--6008.

\bibitem{pix2pix-zero}
G.~Parmar, K.~K. Singh, R.~Zhang, Y.~Li, J.~Lu, and J.~Zhu, ``Zero-shot
  image-to-image translation,'' in \emph{Proceedings of the {ACM} Conference on
  SIGGRAPH}, 2023, pp. 11:1--11:11.

\bibitem{Masactrl}
M.~Cao, X.~Wang, Z.~Qi, Y.~Shan, X.~Qie, and Y.~Zheng, ``Masactrl: Tuning-free
  mutual self-attention control for consistent image synthesis and editing,''
  in \emph{Proceedings of the {IEEE/CVF} International Conference on Computer
  Vision}, 2023, pp. 22\,560--22\,570.

\bibitem{Shape_guided_diffusion}
D.~H. Park, G.~Luo, C.~Toste, S.~Azadi, X.~Liu, M.~Karalashvili, A.~Rohrbach,
  and T.~Darrell, ``Shape-guided diffusion with inside-outside attention,'' in
  \emph{Proceedings of the {IEEE/CVF} Winter Conference on Applications of
  Computer Vision}, 2024, pp. 4198--4207.

\bibitem{ttfdiffusion}
Z.~Yu, J.~Jin, J.~Zhao, Z.~Fu, and J.~Yang, ``Ttfdiffusion: Training-free and
  text-free image editing in diffusion models with structural and semantic
  disentanglement,'' \emph{Neurocomputing}, vol. 619, p. 129159, 2025.

\bibitem{Training-Free_CS}
L.~Schaerf, A.~Alfarano, F.~Silvestri, and L.~Impett, ``Training-free style and
  content transfer by leveraging u-net skip connections in stable diffusion,''
  \emph{CoRR}, vol. abs/2501.14524, 2025.

\bibitem{Blended_diffusion}
O.~Avrahami, D.~Lischinski, and O.~Fried, ``Blended diffusion for text-driven
  editing of natural images,'' in \emph{Proceedings of the {IEEE/CVF}
  Conference on Computer Vision and Pattern Recognition}, 2022, pp.
  18\,208--18\,218.

\bibitem{Blended_latent_diffusion}
O.~Avrahami, O.~Fried, and D.~Lischinski, ``Blended latent diffusion,''
  \emph{ACM Transactions on Graphics}, vol.~42, no.~4, pp. 149:1--149:11, 2023.

\bibitem{Diffedit}
G.~Couairon, J.~Verbeek, H.~Schwenk, and M.~Cord, ``Diffedit: Diffusion-based
  semantic image editing with mask guidance,'' in \emph{Proceedings of the
  International Conference on Learning Representations}, 2023.

\bibitem{Zone}
S.~Li, B.~Zeng, Y.~Feng, S.~Gao, X.~Liu, J.~Liu, L.~Li, X.~Tang, Y.~Hu, J.~Liu
  \emph{et~al.}, ``Zone: Zero-shot instruction-guided local editing,'' in
  \emph{Proceedings of the {IEEE/CVF} Conference on Computer Vision and Pattern
  Recognition}, 2024, pp. 6254--6263.

\bibitem{SAM}
A.~Kirillov, E.~Mintun, N.~Ravi, H.~Mao, C.~Rolland, L.~Gustafson, T.~Xiao,
  S.~Whitehead, A.~C. Berg, W.-Y. Lo \emph{et~al.}, ``Segment anything,'' in
  \emph{Proceedings of the {IEEE/CVF} International Conference on Computer
  Vision}, 2023, pp. 4015--4026.

\bibitem{FISEdit}
Z.~Yu, H.~Li, F.~Fu, X.~Miao, and B.~Cui, ``Accelerating text-to-image editing
  via cache-enabled sparse diffusion inference,'' in \emph{Proceedings of the
  Conference on Association for the Advancement of Artificial Intelligence},
  2024, pp. 16\,605--16\,613.

\bibitem{image_psd_1}
A.~Turiel and N.~Parga, ``The multifractal structure of contrast changes in
  natural images: from sharp edges to textures,'' \emph{Neural computation},
  vol.~12, no.~4, pp. 763--793, 2000.

\bibitem{image_psd_2}
D.~J. Field, ``Relations between the statistics of natural images and the
  response properties of cortical cells,'' \emph{Journal of the Optical Society
  of America A}, vol.~4, no.~12, pp. 2379--2394, 1987.

\bibitem{fft}
K.~R. Castleman, \emph{Digital image processing}.\hskip 1em plus 0.5em minus
  0.4em\relax Prentice Hall Professional Technical Reference, 1979.

\bibitem{dctdiff}
M.~Ning, M.~Li, J.~Su, H.~Jia, L.~Liu, M.~Bene{\v{s}}, W.~Chen, A.~A. Salah,
  and I.~O. Ertugrul, ``Dctdiff: Intriguing properties of image generative
  modeling in the dct space,'' \emph{arXiv preprint arXiv:2412.15032}, 2024.

\bibitem{random_book}
A.~Papoulis, \emph{Random variables and stochastic processes}.\hskip 1em plus
  0.5em minus 0.4em\relax McGraw Hill, 1965.

\bibitem{PIE1}
X.~Ju, A.~Zeng, Y.~Bian, S.~Liu, and Q.~Xu, ``Pnp inversion: Boosting
  diffusion-based editing with 3 lines of code,'' in \emph{Proceedings of the
  International Conference on Learning Representations}, 2024.

\bibitem{h-edit}
T.~Nguyen, K.~Do, D.~Kieu, and T.~Nguyen, ``h-edit: Effective and flexible
  diffusion-based editing via doob’s h-transform,'' in \emph{{Proceedings of
  the {IEEE/CVF} Conference on Computer Vision and Pattern Recognition}}, 2025,
  pp. 28\,490--28\,501.

\bibitem{dino_self}
N.~Tumanyan, O.~Bar-Tal, S.~Bagon, and T.~Dekel, ``Splicing vit features for
  semantic appearance transfer,'' in \emph{Proceedings of the {IEEE/CVF}
  Conference on Computer Vision and Pattern Recognition}, 2022, pp.
  10\,748--10\,757.

\bibitem{lpips}
R.~Zhang, P.~Isola, A.~A. Efros, E.~Shechtman, and O.~Wang, ``The unreasonable
  effectiveness of deep features as a perceptual metric,'' in \emph{Proceedings
  of the {IEEE/CVF} Conference on Computer Vision and Pattern Recognition},
  2018, pp. 586--595.

\bibitem{PSNR}
F.~A. Fardo, V.~H. Conforto, F.~C. De~Oliveira, and P.~S. Rodrigues, ``A formal
  evaluation of psnr as quality measurement parameter for image segmentation
  algorithms,'' \emph{arXiv preprint arXiv:1605.07116}, 2016.

\bibitem{SSIM}
Z.~Wang, A.~Bovik, H.~Sheikh, and E.~Simoncelli, ``Image quality assessment:
  from error visibility to structural similarity,'' \emph{IEEE Transactions on
  Image Processing}, vol.~13, no.~4, pp. 600--612, 2004.

\bibitem{DDIM}
J.~Song, C.~Meng, and S.~Ermon, ``Denoising diffusion implicit models,'' in
  \emph{Proceedings of the International Conference on Learning
  Representations}, 2021.

\bibitem{Classifier-Free}
S.~Sadat, M.~Kansy, O.~Hilliges, and R.~M. Weber, ``No training, no problem:
  Rethinking classifier-free guidance for diffusion models,'' in
  \emph{Proceedings of the International Conference on Learning
  Representations}, 2025.

\bibitem{MS-SSIM}
Z.~Wang, E.~P. Simoncelli, and A.~C. Bovik, ``Multiscale structural similarity
  for image quality assessment,'' in \emph{Proceedings of the {IEEE} Conference
  on Signals, Systems and Computers}, 2003, pp. 1398--1402.

\end{thebibliography}
\bibliographystyle{IEEEtran}
\begin{IEEEbiography}[{\includegraphics[width=1in,height=1.25in,clip,keepaspectratio]{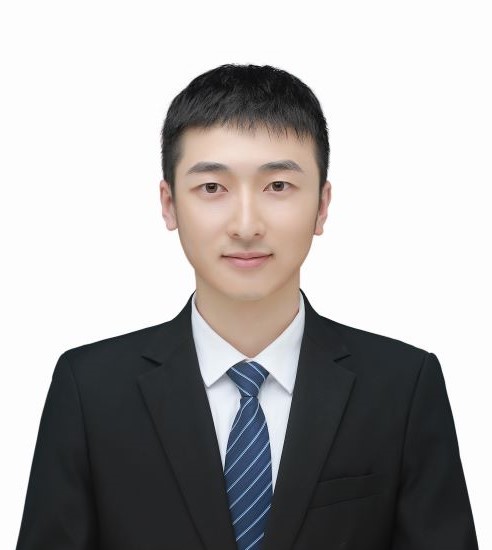}}]{Tiandong Shi} received a master's degree in mining engineering from Central South University, Changsha, China, in 2022. He is currently studying at the School of Geosciences and Info-Physics, Central South University. His current research interests is deep generative model.
\end{IEEEbiography}
\begin{IEEEbiography}[{\includegraphics[width=1in,height=1.25in,clip,keepaspectratio]{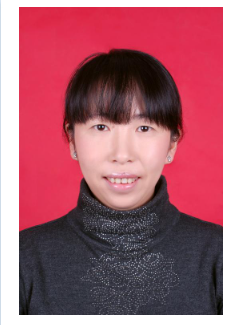}}]{Ling Zhao} received a master's degree in Geodesy and Surveying Engineering from Central South University, Changsha, China, in 2003, and a Ph.D. degree in Cartography and Geographic Information Engineering from Central South University, Changsha, China, in 2013. She is currently a professor at the School of Geosciences and Info-Physics, Central South University, Changsha, China. Her current research interests include geo/remote sensing big data, machine/deep learning.
\end{IEEEbiography}
\begin{IEEEbiography}[{\includegraphics[width=1in,height=1.25in,clip,keepaspectratio]{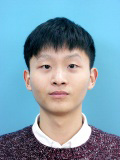}}]{Ji Qi} received a B.S. degree in remote sensing science and technology from Central South University, Changsha, China, in 2018, and Ph.D. degree in surveying and mapping science and technology from Central South University, Changsha, China, in 2024. He is currently a postdoctoral with the School of Geography and Remote Sensing, Guangzhou University, Guangzhou, China. His research interests include computer vision, continual learning, and remote sensing image processing.
\end{IEEEbiography}
\begin{IEEEbiography}[{\includegraphics[width=1in,height=1.25in,clip,keepaspectratio]{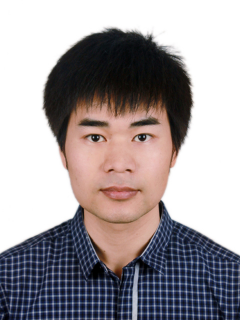}}]{Jiayi Ma}
(M'14-SM'21) received the B.S. degree in information and computing science and the Ph.D. degree in control science and engineering from the Huazhong University of Science and Technology, Wuhan, China, in 2008 and 2014, respectively. He is currently a Professor with the Electronic Information School, and also with the School of Robotics, Wuhan University, Wuhan, China. He has coauthored more than 400 refereed journal and conference papers, including Cell, IEEE TPAMI, IJCV, etc. He is a recipient of the IEEE SPS Best Paper Award 2025, the \emph{Information Fusion} Best Paper Award 2024, and the Hsue-shen Tsien Paper Award 2023. He is a Co Editor-in-Chief of \emph{Information Fusion}, an Associate Editor of \emph{IEEE Transactions on Image Processing} and \emph{IEEE/CAA Journal of Automatica Sinica}, and a Youth Editor of \emph{The Innovation} and \emph{Fundamental Research}.
\end{IEEEbiography}
\begin{IEEEbiography}[{\includegraphics[width=1in,height=1.25in,clip,keepaspectratio]{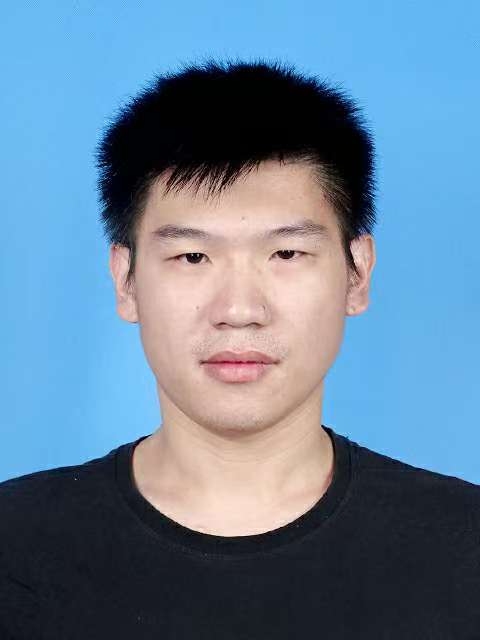}}]{Chengli Peng} received the B.E. degree from the School of Electrical Engineering, Xinjiang University, Urumchi, China, in 2016, the M.S. degree from the School of Engineering, Huazhong Agricultural University, Wuhan, China, in 2018, and the Ph.D. degree from the Electronic Information School, Wuhan University, Wuhan, China, in 2021. He is currently a Research Associate with the School of Geosciences and Info-Physics, Central South University, Changsha, China. His research interests include computer vision and deep learning.
\end{IEEEbiography}

\end{document}